%% file: main.tex
\documentclass[10pt,twocolumn,letterpaper]{article}

\usepackage[pagenumbers]{iccv} 
\usepackage{multicol}
\usepackage{multirow}
\usepackage{cuted}
\usepackage{subfiles}
\usepackage{algorithmicx}
\algrenewcommand{\algorithmiccomment}[1]{\hskip3em$\rightarrow$ #1}

\usepackage{stfloats}
\usepackage{arydshln}
\input{preamble}

\definecolor{iccvblue}{rgb}{0.21,0.49,0.74}
\usepackage[pagebackref,breaklinks,colorlinks,allcolors=iccvblue]{hyperref}

\def\paperID{3438} 
\def\confName{ICCV}
\def\confYear{2025}

\title{DynamicFace: High-Quality and Consistent Face Swapping for Image and Video using Composable 3D Facial Priors}

\author{
  Runqi Wang\textsuperscript{1,2} \ \ 
  Yang Chen\textsuperscript{1} \ \ 
  Sijie Xu\textsuperscript{1} \ \ 
  Tianyao He\textsuperscript{1,3} \ \ 
  Wei Zhu\textsuperscript{1} \ \ 
  Dejia Song\textsuperscript{1} \\
  Nemo Chen\textsuperscript{1} \ \ 
  Xu Tang\textsuperscript{1} \ \ 
  Yao Hu\textsuperscript{1} \\
  \textsuperscript{1}Xiaohongshu \quad
  \textsuperscript{2}ShanghaiTech University \quad
  \textsuperscript{3}Shanghai Jiao Tong University \\
}

\begin{document}
\maketitle
\begin{strip}
    \centering
    \includegraphics[width=1.0\linewidth]{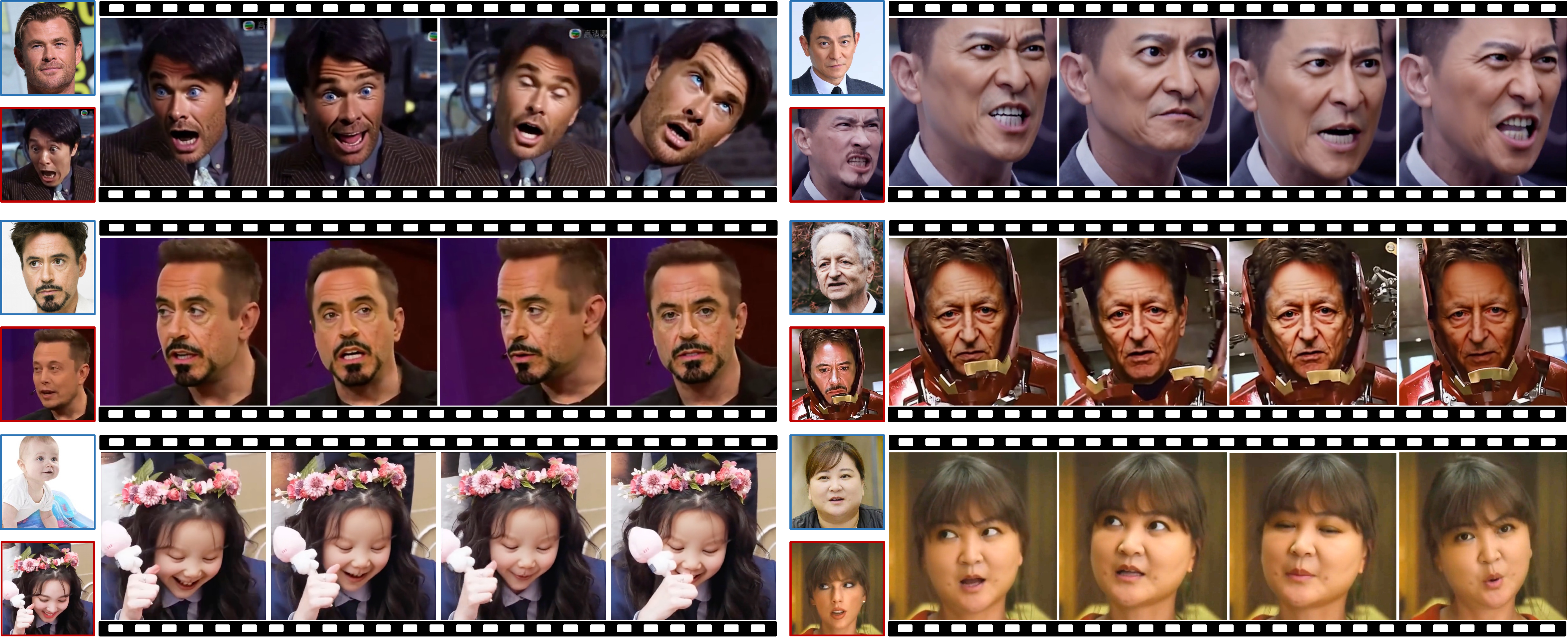}
    \captionof{figure}{\textbf{Video face swapping.} We adopt precise conditions and motion module for better temporal consistency. Given source face image and target face frames, DynamicFace can generate faces with high identity consistency and motion accuracy.}
    \label{fig:video_showcase}
\end{strip}
\input{sec/0_abstract}    
\input{sec/1_intro}
\input{sec/2_relatedwork}
\input{sec/3_method}
\input{sec/4_experiments}
\input{sec/5_conclusion}
{
    \small
    \bibliographystyle{ieeenat_fullname}
    \bibliography{main}
}

\clearpage
\appendix

\begin{strip}
  \centering
  \Large \textbf{DynamicFace: High-Quality and Consistent Face Swapping for Image and Video using Composable 3D Facial Priors} \\
  \vspace{2pt}
  Supplementary Materials
  \vspace{1em}

\end{strip}

\subfile{supp}

\end{document}

%% file: preamble.tex
\usepackage{adjustbox}
\usepackage{algorithm}
\usepackage{subcaption}
\usepackage{algorithmicx}
\usepackage{amsmath}
\usepackage{amssymb}
\usepackage{arydshln}
\usepackage{array}
\usepackage{caption}
\usepackage{color}
\usepackage{enumitem}
\usepackage{epsfig}
\usepackage{epstopdf}
\usepackage{flushend}
\usepackage{graphicx}
\usepackage{makecell}
\usepackage{multirow}
\usepackage{multicol}
\definecolor{ngreen}{RGB}{17, 173, 30}
\usepackage{soul}
\usepackage{tabularx}
\usepackage{times}
\usepackage{utfsym}
\usepackage[normalem]{ulem}
\usepackage{xcolor}
\usepackage{bm}
\usepackage{algpseudocode}

%% file: sec/0_abstract.tex
\begin{abstract}


Face swapping transfers the identity of a source face to a target face while retaining the attributes like expression, pose, hair, and background of the target face. Advanced face swapping methods have achieved attractive results. However, these methods often inadvertently transfer identity information from the target face, compromising expression-related details and accurate identity. We propose a novel method DynamicFace that leverages the power of diffusion models and plug-and-play adaptive attention layers for image and video face swapping. First, we introduce four fine-grained facial conditions using 3D facial priors. All conditions are designed to be disentangled from each other for precise and unique control. Then, we adopt Face Former and ReferenceNet for high-level and detailed identity injection. Through experiments on the FF++ dataset, we demonstrate that our method achieves state-of-the-art results in face swapping, showcasing superior image quality, identity preservation, and expression accuracy. Our framework seamlessly adapts to both image and video domains. Our code and results will be available on the project page: \url{https://dynamic-face.github.io/}.
\end{abstract}

%% file: sec/1_intro.tex
\section{Introduction}
\label{sec:intro}

Face swapping has attracted many interests because of its wide applications, such as portrait reenactment, film production, and virtual reality. But there are still two pivotal challenges: 1) balance identity from source face and motion from target face. 2) give precise non-identity motion guidance to maintain temporal consistency. 

Recent works~\cite{li2019faceshifter,deepfakes,shen2018neural} have made great efforts to achieve good face swapping results. However, these methods often focus on inner facial texture but ignore shape and illumination. Later, some works~\cite{wang2021hififace,xu2021facecontroller,mi2025data} combine the expert knowledge from 3D face reconstruction to achieve better performance. But training model with both reconstruction loss and identity loss will face a trade-off problem between identity and reconstruction, which often leads to instability and unreliability. Balancing identity and reconstruction with only losses constraint is conflict. We should provide accurate information for controllable generation.

Recently, diffusion-based models~\cite{ho2020denoising,song2021scorebased,rombach2022ldm,he2025fulldit2} have exhibited high customizability for various conditions and impressive ability to generate images with high resolution and complex scenes. Some works~\cite{han2024faceadapter,zhao2023diffswap,kim2022diffface} try to utilize the diversity and powerful generation ability of diffusion model to achieve face swapping. However, these methods could not achieve video face swapping directly. And idealized face guidance should preserve the non-identity attributes of $I_{tgt}$, which include non-facial attributes (e.g., background and hair), facial posture (e.g., expression and pose) and facial color (e.g., lighting). Previous works didn't pay much attention to the potentiality of 3D facial priors to simultaneously maintain identity from source face and complicated motion information from target face.

In this work, we propose a method named DynamicFace for face swapping by incorporating precise and disentangled facial conditions to the powerful Stable Diffusion model~\cite{rombach2022ldm} and apply temporal attention layer in AnimateDiff~\cite{guo2024animatediff} to make DynamicFace also available in video domain. We introduce a new multiple conditioning mechanism to address the limitations of previous methods. Specifically, 1) to accurately control expression, pose, and other motion, we introduce four fine-grained facial conditions: background, shape-aware normal map, expression-related landmark, and identity-removed UV texture map. These four conditions fully encapsulate the desired information in the target face and are well disentangled from each other. 2) to avoid introducing unexpected identity information from the target face, we propose a alignment strategy to form spatial-aligned conditions using 3D facial priors. This involves extracting pose and expression parameters from the target face video and shape parameters from the source image using the prior of 3D face reconstruction model 3DDFA-V3~\cite{wang2024_3ddfav3}, and then rendering the 2D face normal map and the UV texture map with these facial parameters. In summary, our main contributions are as follows:

\begin{figure}[t]
    \centering
    \includegraphics[width=\linewidth]{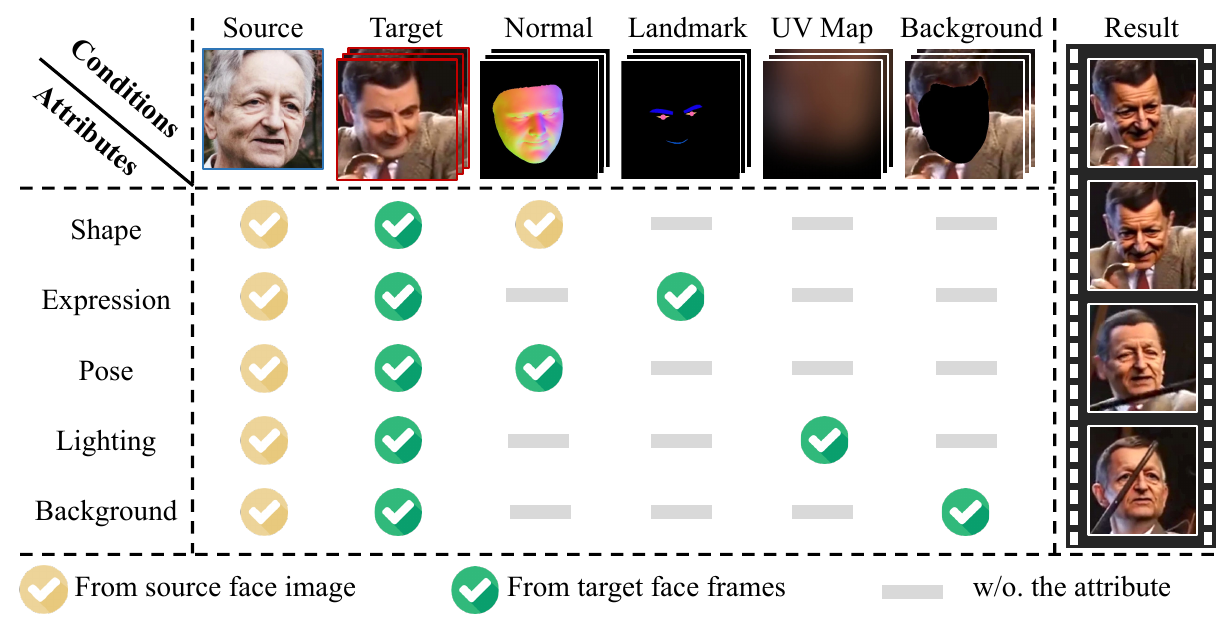}

    \caption{\textbf{Composable face conditions.} We aim to decompose face into four conditions and capture its unique usage of each condition. Conditions are disentangled with each other and provide essential guidance with 3D facial priors.}
    \label{fig:introduction}

\end{figure}

\begin{itemize}
    \item We propose DynamicFace, a novel diffusion-based video face swapping approach that could generate high-fidelity and consistent faces leveraging the prior knowledge of Stable Diffusion.
    \item To introduce precise guidance when controlling face motion, we decompose the face into four conditions: background, shape-aware normal map, expression-related landmark, and identity-removed UV texture map. We disentangle conditions with each other using 3D facial priors for its distinct usage, which preserves shape aligned with source image and non-identity attributes.
    \item We  propose the Fusion TV Optimizer (FusionTVO), a plug-and-play inference module that innovatively combines position-weighted latent fusion and temporal TV regularization to eliminate segment flicker artifacts.
\end{itemize}

%% file: sec/2_relatedwork.tex
\section{Related Work}
\label{sec:formatting}

Recently, face swapping has drawn much attention from the research community, and it has many applications in visual effects. Face swapping means transferring the identity information of the source image to the target image while keeping the other attributes like the expression and background of the target image unchanged.



\subsection{Face Swapping}
Early methods directly warp the source face according to the target facial landmarks, thus failing to address the large differences in posture and expression. 3DMM-based methods swap faces by 3D-fitting and re-rendering. However, these methods often cannot handle lighting differences on face and suffer from poor fidelity. Later, GAN-based methods improve the fidelity of the generated faces. Deepfakes~\cite{deepfakes} transfers the target attributes to the source face by an encoder-decoder structure while being constrained by two specific identities.
FSGAN~\cite{nirkin2019fsgan} animates the source face with target facial landmarks and fuses it with the target background. SimSwap~\cite{chen2020simswap} introduces a feature matching loss to help preserve the target attributes. 
These methods require a facial mask for blending, which limits face shape variation.

Diffusion models have shown remarkable performance in various generative tasks, emerging as a powerful alternative to traditional GANs for image and video synthesis. The innovative approach of these models lies in their ability to gradually refine an image by reversing a learned noising process, which has led to significant advancements in image quality and diversity. In the context of face swapping, diffusion models have introduced a new paradigm for identity manipulation in face images. DiffSwap~\cite{zhao2023diffswap} proposes a diffusion-based pipeline in latent space for face swapping by integrating 3D-aware masked diffusion and midpoint estimation during the reverse process. DiffFace~\cite{kim2022diffface} introduces an ID conditional DDPM. However, pure pixel-wise diffusion without pretraining requires more data, and extra forward steps with gradients make training more memory-consuming. These both make training process difficult. Face Adapter~\cite{han2024faceadapter} only utilizes an efficient adapter to introduce identity information and preserve attributes of the target face image, but it is difficult to achieve temporal consistency when balancing the trade-off between identity and motion. In addition, these diffusion-based methods could not easily transfer to the video domain.

\begin{figure*}[ht]
    \centering
    \includegraphics[width=1.0\linewidth]{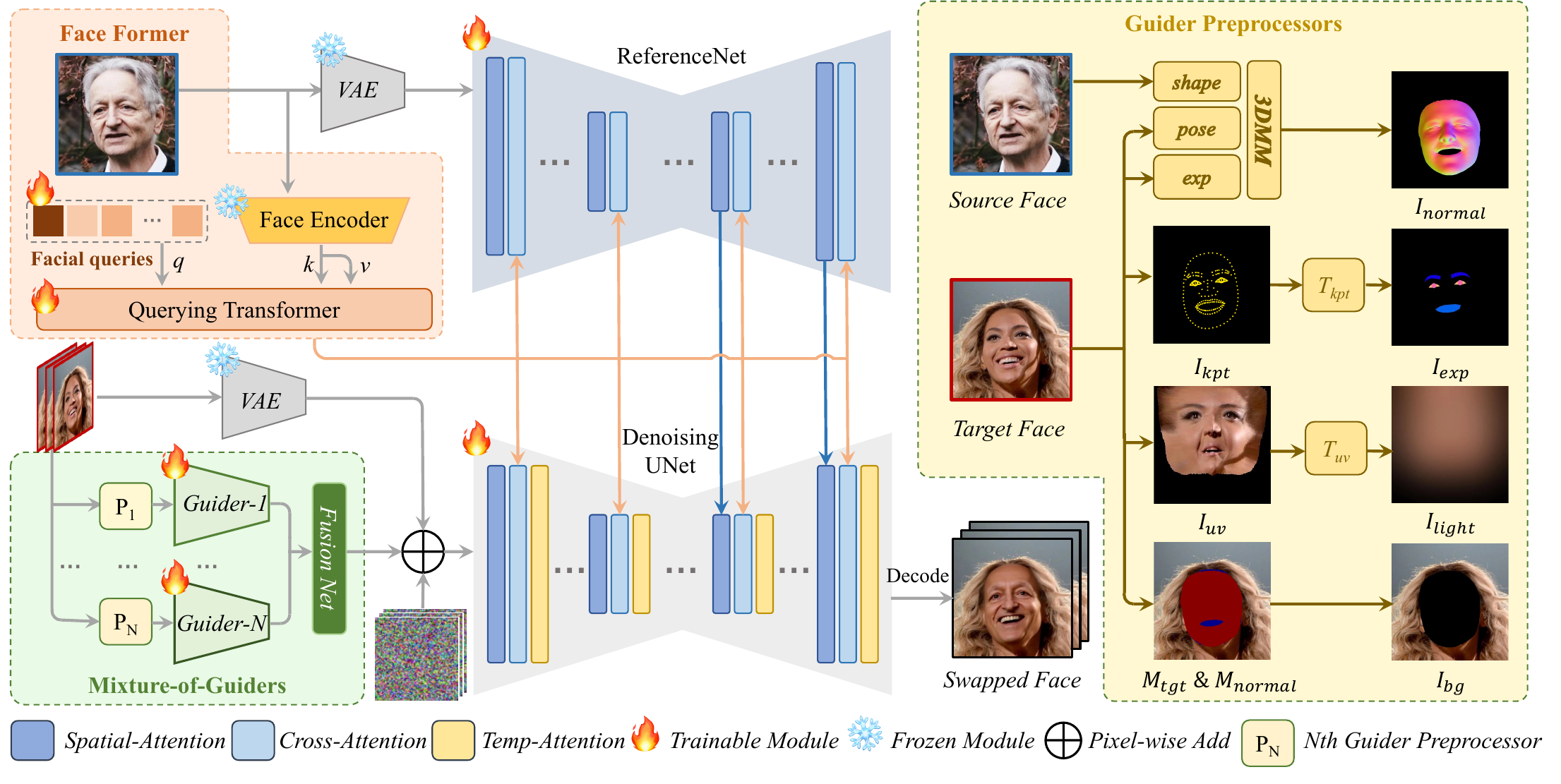}
    \caption{\textbf{The overview of the proposed method.} A VAE encoder and the ReferenceNet extract detailed features from the source face, which are then merged into the Stable Diffusion main UNet via spatial-attention. The face encoder~\cite{deng2019arcface} extracts high-level features from the source face image with querying transformer, which is then injected into both the ReferenceNet and the main UNet via cross-attention. Four composable face conditions are fed into four expert guiders and fused with fusion net in the latent space. Temporal-attention aims to improve the temporal consistency across frames. After iterative denoising, the output of the main UNet is decoded into the final animated video by a VAE decoder.}
    \label{fig:enter-label}
\end{figure*}

\subsection{Identity-Preserving Generation}
The impressive generative abilities of foundation models have attracted recent research endeavors investigating their personalized generation potential. With popular pretrained T2I and T2V diffusion models like Stable Diffusion~\cite{rombach2022ldm} and AnimateDiff~\cite{guo2024animatediff}, several ID-Preserving methods are proposed with promising results. IP-Adapter~\cite{ye2023ip} and InstantID~\cite{wang2024instantid} could maintain high face fidelity with effective adaptable plugins. But, it's hard to accurately control expression and pose without rich motion guidance, which makes video generation difficult. Animate Anyone~\cite{hu2024animate} intends to use spatial-attention to extract identity from source face image and pose sequence to offer motion information, but this lacks appearance alignment which results in inaccurate identity. Champ~\cite{zhu2024champ} aims to introduce multiple pose conditions from 3D human parametric model for accurately capturing both pose and shape variations. However, the conditions of the champ are not sufficiently disentangled, leading to redundant information. In this paper, we aim to first decompose all facial conditions with their own unique usage and to make full use of the prior knowledge of Stable Diffusion for face swapping.

%% file: sec/3_method.tex
\section{Method}
We propose a method to empower the diffusion model with motion-controllable modules and identity-injection modules, allowing for video face swapping with one source face image and target face frames. In this section, we will first introduce the brief prior knowledge of latent diffusion models. Then, we give the details of composable face conditions and Mixture-of-Guiders for controlling motion and appearance-aware modules to accurately preserve identity.
\subsection{Preliminary: Latent Diffusion Models}
Diffusion models generate a realistic image from a standard Gaussian distribution by reversing a recurrent noising process~\cite{ho2020denoising}.  However, pure diffusion process in pixel-space is really memory-consuming. Latent Diffusion Model~\cite{rombach2022ldm} makes the forward and reverse process in the latent space to reduce computation with an autoencoder. 
It first encodes the desired image into lower-dimension: $\mathbf{z}=\mathcal{E}(x)$. Then, LDM applies diffusion in the latent space. The forward process gradually alters to Gaussian distribution from the data $\mathbf{z}_0 \sim q(\mathbf{z}_0)$, which could be concluded as $q(\mathbf{z}_t|\mathbf{z}_{0}):=\mathcal{N}(\mathbf{z}_{t};\sqrt{\bar\alpha_t}\mathbf{z}_{t-1}, (1-\bar\alpha_t) \mathbf{I})$, where $\bar\alpha_t$ is a predefined mean schedule.
In addition, the reverse process is as follows:
\begin{equation} \label{eq:forward_ddpm}
    p_\theta(\mathbf{z}_{t-1}|\mathbf{z}_t):=\mathcal{N}(\mathbf{z}_{t-1}; \mu_\theta (\mathbf{z}_t, t), \sigma_\theta(\mathbf{z}_t, t)\mathbf{I}),
\end{equation}
where $\mu_\theta$ and $\sigma_\theta$ are parameterized with neural network. A common training strategy that uses noise approximation model $\epsilon_\theta(\mathbf{z}_t, t,c)$ performs better than other forms from experiments in previous work~\cite{nichol2021improved}. The training objective could be represented by a simplified L2 loss between ground truth noise and predicted noise:
\begin{equation}
\mathcal{L}_{ldm}=\mathbb{E}_\theta\left[\left\|\boldsymbol{\epsilon}-\boldsymbol{\epsilon}_\theta\left(\mathbf{z}_t, t, c\right)\right\|^2\right],
\label{eq:ldmloss}
\end{equation}
where $c$ is facial conditions, and $t$ represents the timestep of current noisy image. At last, we use the decoder to reconstruct image from the latent output: $\hat{x}=\mathcal{D}(\mathbf{z}_0)$.

\subsection{Composable Face Conditions}
The goal of face swapping is to preserve the identity of source face image and spatially align well with the target face image. Given a source face image $I_{src}$ and a sequence of target face video frames $I_{tgt}^{1:N}$, we first adopt a 3D facial reconstruction method to obtain 3D face model, $F_{src}$ and $F^{1:N}_{tgt}$, containing different geometry parameters like identity $\alpha$, pose $\beta$, expression $\theta$ and albedo $\gamma$. We intend to combine the identity of source face and motion of target face together. Figure~\ref{fig:introduction} illustrates the attribute information contained in each condition and its corresponding source. The key of the composable face condition in our work is to design more accurate and independent condition for flexible and accurate controlling. To make conditions disentangled with each other, we decompose faces into four conditions: background, expression, shape, and illumination. The details are as follows:
    \noindent\textbf{Shape-Aware Pose Condition.} We use the 3DDFA-V3~\cite{wang2024_3ddfav3} to estimate facial parameters from each frame of  target face video and source face image. Then, we render the 2D normal map from 3D face for pose and shape information. However, apart from the necessary pose and expression details, these images also encapsulate identity-specific facial geometry from the target individuals. The identity information essential for our purpose should be extracted from the source image. To address this mismatching issue, we utilize a facial shape alignment strategy which combines the identity information from the source image with the pose and expression information from the driving frames to generate the refined 2D facial conditions:
    \begin{equation}
    I_{normal}^{1:N} = \mathcal{R}_{normal}(F(\alpha_{src}, \beta^{1:N}_{tgt}, \theta^{1:N}_{base}), c^{1:N}_{tgt}),
    \end{equation}
    where $I_{normal}^{1:N}$ represents the shape and pose information of 2D facial conditions, the rendered surface normal map, for frames $1:N$. $\mathcal{R}_{normal}$ represents the rendering operation with normal texture of the mesh. $F(\cdot)$ stands for 3D face model used to obtain face vertices with facial parameters. $c$ represents the camera information to project the 3D mesh into image space. $\alpha_{src}$ is extracted from the source face image, while $\beta^{1:N}_{tgt}$, and $c^{1:N}_{tgt}$ are pose, camera parameters extracted from the target face frames and $\theta^{1:N}_{base}$ is expression parameter from template face. We intend to remove specific expression information from target face to better disentangle expression condition and pose condition.
    
    \noindent\textbf{Background-Preservation Condition.} 
    We only need to change facial area in face swapping task. So, the background information could be directly sent to the model as a conditional inpainting task. The key of designing background condition is how to generate face mask. Most face swapping methods use the original facial mask of target face image, but this will ignore the face shape differences during inference. Face Adapter~\cite{han2024faceadapter} uses dilated facial area to train an area predictor for possible inpainting mask esstimation. Nevertheless, finding an optimal dilation size is crucial, as a face mask with small dilation also introduces useless shape information of the target face image, whereas a face mask with large dilation may contain face occlusion that makes model training more difficult. Here, we first use a pretrained face parser~\cite{yu2018bisenet} to predict facial area $M_{tgt}$ and occlusion area $M_{occ}$. To avoid learning facial shape of target face during training, we apply random shift on the facial area $M_{tgt}$ to get $M_{tgt}^{shift}$. To address the shape misalignment problem in the inference phase, we replace $M_{tgt}^{shift}$ with $M_{tgt}^{normal}$ from the area of $I_{normal}$, containing shape of source face and position of target face. The swapping area $M_{swap}$ could be described as follow:

    \begin{align}
    M_{swap} = \begin{cases}
        (M_{tgt} \cup M_{tgt}^{shift}) \cap  (1 - M_{occ}) &\textit{Train}\\[4pt]
       (M_{tgt} \cup M_{tgt}^{normal}) \cap  (1 - M_{occ})&\textit{Test}
    \end{cases}\label{eq:param1}
    \end{align}
    Finally, we can decompose non-indentity background condition from target face image for environmental information injection:
    \begin{equation}
            I_{bg}^{1:N} = (1 - M_{swap}^{1:N}) \circ I_{tgt}^{1:N}
    \end{equation}

    \noindent\textbf{Expression-Related Condition.}
    Facial expression is essential for face swapping. It's hard to combine identity of source face image and facial expression of target face image. Some methods~\cite{zhao2023diffswap, han2024faceadapter} use 3D face reconstruction to get 3D-aware keypoints by replacing the shape with source face image. However, the 3D-aware keypoints miss the detailed information of the eyes and its precision is suboptimal. In this work, we adopt 2D landmark of target face image, and transform points to segmentation for a simple and straight representation. We use several sets of boundary landmark to generate expression segmentation including areas of eyebrows, eyes, eyeballs and mouth:
    \begin{equation}
        I_{exp}^{1:N} = H(l_{eyebrows}^{1:N}, l_{eyes}^{1:N}, l_{eyeballs}^{1:N}, l_{mouth}^{1:N})
    \end{equation}
    where $H(\cdot)$ represents that we compute several convex hulls of given landmarks and combine these semantic areas together. We strive to remove identity information of target face image, for example, only use the area inside mouth to prevent bringing the thickness of mouth which may include identity.
    
    \noindent\textbf{Identity-Erased Illumination Condition.}
    Environmental lighting is also a necessary condition to make generated face frames well aligned with target face frames. Few methods pay attention to illumination in face swapping task. With the 3D facial prior in previous section, we could use the UV texture map of target face frames to provide illumination information. UV texture map contains irrelevant identity of target face image, while we only need to capture lighting from it. So, we add blur on the rendered UV texture map to destroy target face identity as:

    \begin{equation}
    I_{light}^{1:N} = Blur(\mathcal{R}_{uv}(V_{3d}, Tex_{uv})),
    \end{equation}
    where $I_{light}^{1:N}$ means the illumination condition with 2D representation, for frames $1:N$. $\mathcal{R}_{uv}$ stands for the rendering operation with UV mapping. $V_{3d}$ represents vertices of 3D face mesh and $Tex_{uv}$ stands for the texture color of UV mapping. More details are shown in ablation study.
\subsection{Mixture-of-Guiders}
Now, we have completed four composable face conditions with their unique usage by leveraging the 3D facial prior of both source face image and target face frames. All conditions are spatially mapped into image-level motion sequences. Inspired by the previous work~\cite{zhu2024champ}, we use Mixture-of-Guiders for lightweight guidance instead of several ControlNets. All guiders are implemented as a set of convolution and activation layers, aiming to extract local spatial information. This architecture facilitates condition adaptability while promoting parameter efficiency. We also add a self-attention module to learn the unique information from each disentangled face condition. This could capture more global information from each guidance. Even though all conditions are well designed for its distinct character and spatially aligned with generated face, there are still some mismatching which could not be solved due to limitation of 3D reconstruction and compression when passing to guiders. The conditions from some works are not disentangled from each other, which contains too much redundant information, and they directly fuse these guidance condition through summation. This would cause the network only learn useful information from some detailed condition and drop some of repeated condition for easier learning. We adopt a fusion net to better fuse these four distinct face conditions. We first concat all face conditions after each guiders and use a self-attention mechanism to learn the spatial relation among these conditions. Both guiders and fusion net apply zero convolution at the last layer to maintain the prior knowledge of original Stable Diffusion and make the whole architecture more lightweight to train with motion guidance.
\subsection{Appearance-Aware Controlling}
\noindent\textbf{Face Former.}
In this work, to enhance the ability of identity injection, Face Former is proposed to enable the diffusion model to accurately preserve the identity with the face image prompt. In text-to-video tasks, textual prompts primarily ensure semantic relevance with the generated visual content, whereas image-to-video tasks demand precise consistency due to the detailed features encapsulated in images. Prior studies in image-driven generation have used the CLIP image encoder as a substitute for the text encoder in cross-attention layer, but this design struggles with accurate facial representation. Here, we use a face recognition model instead of the vision-language CLIP model~\cite{radford2021clip} to extract face embedding from the face image prompt. To effectively inject the face embedding to the original diffusion model, we use a lightweight trainable querying transformer like recent work~\cite{li2023blip2,han2024faceadapter,he2024idanimator} to transform identity token space to semantic-level textual space. We create $N_{token}$ learnable facial queries as input for the querying transformer. The facial queries will interact with face embedding through each self-attention layer. Thus, these tokens could be sent to the UNet for high-level identity injection in face swapping.

\noindent\textbf{ReferenceNet.}
To preserve dense facial textures and detailed identity information from the source face image, we adopt ReferenceNet for fine-grained identity injection. While Face Former encodes low-resolution inputs (112×112), leading to the loss of subtle features due to its emphasis on high-level feature matching, ReferenceNet addresses this limitation by leveraging a trainable replica of Stable Diffusion's base UNet. Inspired by MasaCtrl~\cite{cao_2023_masactrl} and Animate Anyone~\cite{hu2024animate}, ReferenceNet integrates appearance features through a spatial attention mechanism during denoising. By replacing self-attention layers with spatial-attention modules in intermediate and upsampling stages, it maintains the semantic layout of the source face while preserving texture details and intra-personal attributes, ensuring robust identity consistency.
\begin{figure}[t]
    \centering
    \includegraphics[width=\linewidth]{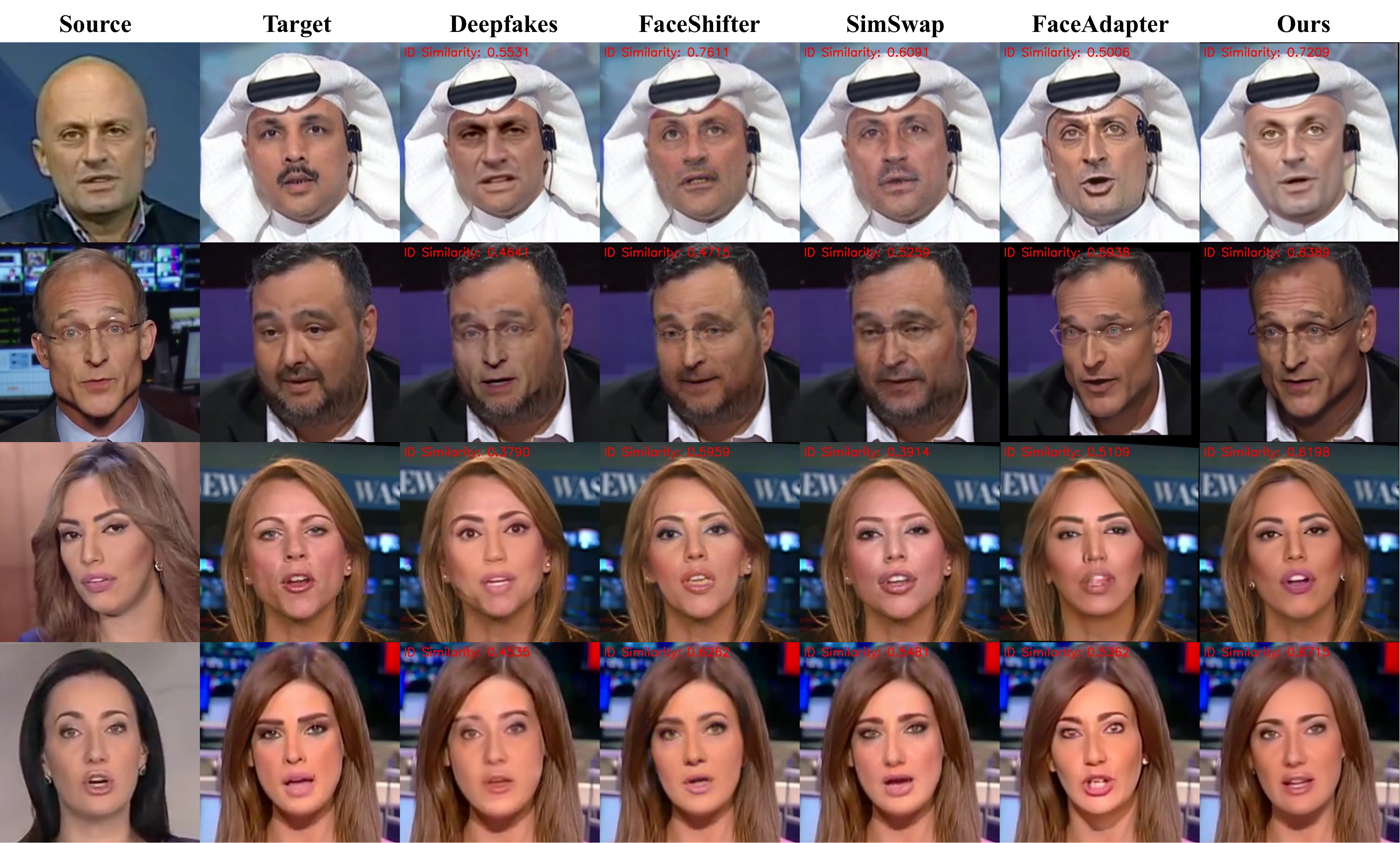}
    \caption{\textbf{Qualitative comparisons on FF++.} Our method performs well to unseen data distribution and can also better preserve both the identity (e.g., shape and facial texture) and the motion including expression and pose.}
    \label{fig:ff_comparison}
\end{figure}

\subsection{Latent Fusion Total Variation Optimizer}

In long-duration video face swapping, abrupt inter-frame motions (e.g., rapid head rotations or expression changes) often lead to temporal flickering and inconsistent identity rendering. To address this, we propose FusionTVO, a plug-and-play module that integrates segment-aware latent fusion and temporal Total Variation (TV) regularization to ensure smooth transitions across frames. While our composable facial conditions and temporal attention layers already ensure coarse temporal alignment, subtle flickering artifacts may persist due to per-frame denoising variations. Given two adjacent segments, overlapping frames near their boundary are assigned fusion weights based on their relative positions. This ensures smooth transitions between segments by prioritizing frames closer to their original segment’s center. To suppress flickering caused by large motions, we impose a TV constraint on consecutive latents without retraining the model. We integrate the TV term into the reverse diffusion process using Half-Quadratic Splitting (HQS) algorithm ~\cite{hqs1995}, decoupling the denoising objective (guided by facial conditions) from the temporal consistency constraint.
\begin{equation}
\mathbf{z}_{t-1}^{(1: N)}=\arg \min _{\mathbf{z}} \underbrace{\left\|\mathbf{z}-\hat{\mathbf{z}}_{t-1}^{(1: N)}\right\|_2^2}_{\text {Denoising Prior }}+\mathcal{R}_{T V}(\mathbf{z}),
\end{equation}
where 
$\mathbf{z}_{t-1}^{(1: N)}$ is the initial latent prediction from the diffusion model. For algorithmic details, including pseudocode and hyperparameter, we refer readers to Algorithm \ref{alg:PnP}.
\begin{algorithm}[t]\caption{FusionTVO}\label{alg:PnP}
\begin{algorithmic}[1]
      \State \textbf{Input}: Segments\{$S_1, S_2,...,S_K$\}, Overlap Size $O$, Facial Conditions $c^{1:N}$, Latent Face Mask $\tilde{M}_{swap}^{1:N}$, Latent Target Video $\mathbf{z}_\text{target}^{1:N}$, ${\{\sigma_{t}\}_{t=1}^T}$, $\mathbf{z}_{T}^{1:N}$, $\lambda_\text{fusion}$, $\lambda_\text{TV}$.
      \State \textbf{Output}: Fused Face Swapping Video $\mathbf{\hat{z}}_{0}^{1:N}$.

      \vspace{1mm}

    \For{$t=T$ {\bfseries to} $1$} 
    
        \For{$n=1$ {\bfseries to} $K$} \Comment{ Denoising steps}
        
         \State{$ \mathbf{\epsilon}_{t} \sim \mathcal{N}(\mathbf{0}, \mathbf{I})$}
         
         \State{$\mathbf{\hat{z}}_\text{target}^{1:N} = \sqrt{\bar\alpha_{t}}\mathbf{z}_\text{target}^{1:N} + (1-\bar\alpha_t) \mathbf{\epsilon}_{t} $}
         \State{$\mathbf{\hat{z}}_0^{S_n} = \frac{1}{\sqrt{\alpha_t}}\big(\mathbf{{z}}_{t}^{S_n} - \frac{\beta_t}{\sqrt{1-\bar{\alpha}_t}}\hat{\epsilon}_{\theta}(\mathbf{{z}}_{t}^{S_n}, c^{S_n}, t)\big) \circ  \tilde{M}_{swap}^{S_n} + \mathbf{\hat{z}}_\text{target}^{1:N} \circ (1-\tilde{M}_{swap}^{S_n}) $}\Comment{Repaint Background} 
      \EndFor
          \For{each adjacent pair $(S_k, S_{k+1})$} 
          \State{Extract overlapping frames $\{\mathbf{{\tilde{z}}}^{(i)}\}_{i=1}^O$ }
              \For{$i=1$ {\bfseries to} $O$}\Comment{Weighted Fusion} 
              \State{$d_i = O - i + 1, w_i = d_i\cdot\lambda_\text{fusion}$ }
              \State{$\mathbf{{\tilde{z}}}^{(i)} = w_i  \mathbf{{\hat{z}}}_{S_k}^{(i)} +(1-w_i)\mathbf{{\hat{z}}}_{S_{k+1}}^{(i)} $  }
              \EndFor
              \State{Update current overalpping frames in $\mathbf{\hat{z}}_{t-1}^{1:N}$}
          \EndFor
      
      \State{$ \mathbf{\hat{z}}_{t-1}^{1:N}=\mathbf{{\hat{z}}}_{t-1}^{1:N} - {\lambda_\text{TV}}\nabla_{\mathbf{\hat{z}}_{t-1}^{1:N}} \|D_{z} \mathbf{\hat{z}}_{t-1}^{1:N}\|^2$}  \Comment{TV Opti.}
    \EndFor
      \State {\bfseries return} $\mathbf{\hat{z}}_{0}^{1:N}$
\end{algorithmic}
\end{algorithm}


%% file: sec/4_experiments.tex
\section{Experiments}

\begin{figure*}[ht]
    \centering
    \includegraphics[width=1.0\linewidth]{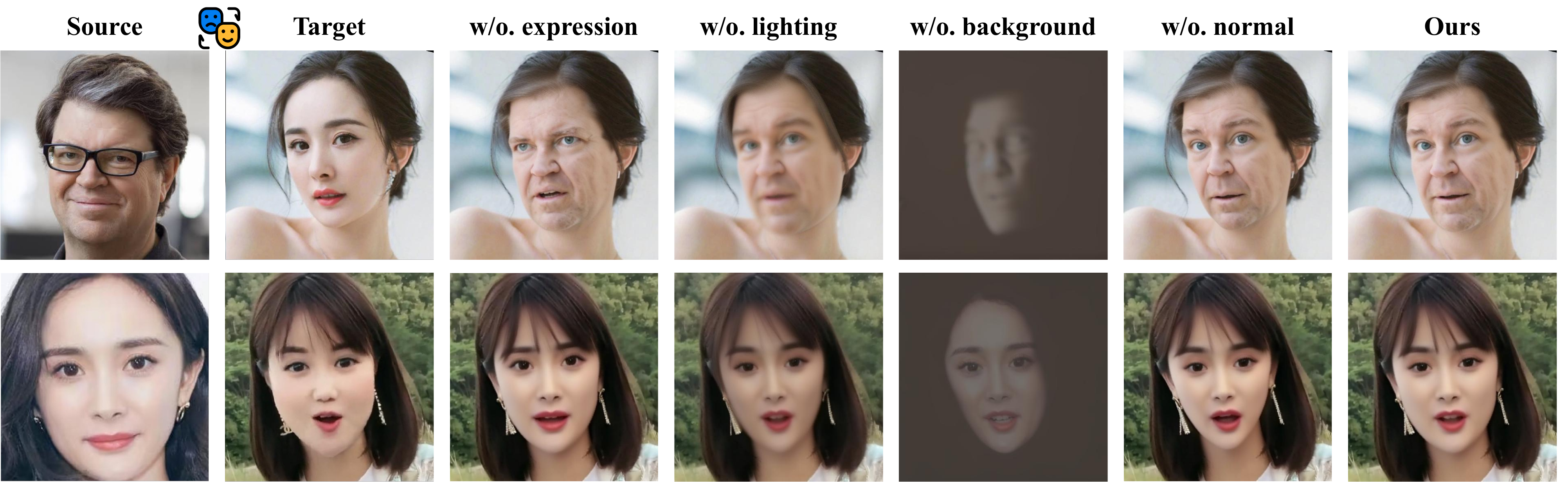}
    \caption{\textbf{Ablation study for different conditions.} We explore the contribution of different facial conditions. Results show that each face guidance has its own unique effect. }
    \label{fig:ablation_figure}
\end{figure*}

\subsection{Implementation Details}

\begin{figure}[h]
    \centering
    \includegraphics[width=\linewidth]{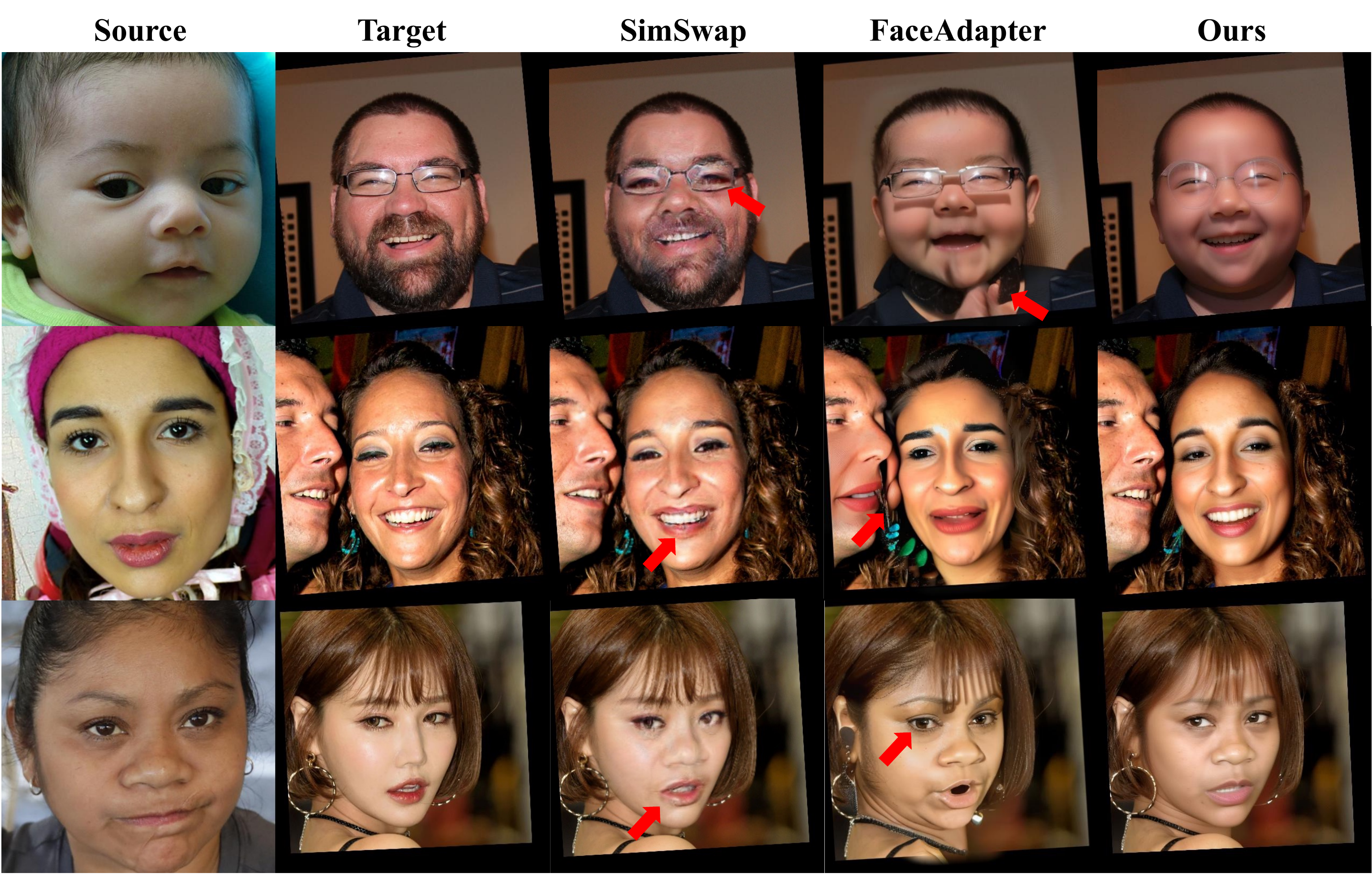}
    \caption{\textbf{Qualitative comparisons on FFHQ.} 
    Our method can generate high-resolution faces with accurate identity and precise motion containing expression, pose and gaze. The background of our results is also more realistic.}
    \label{fig:ffhq_comparison}
\end{figure}
We choose VGGFace2~\cite{cao2018vggface2}, VFHQ~\cite{xie2022vfhq} and a private dataset as the training set. Particularly, low-quality faces are removed to ensure high-quality training. For evaluation, we select FaceForensics++ dataset to test. 
The training process consists of two stages. In the initial training stage, we sample faces from same individuals to train spatial modules including face querying transformer and ReferenceNet for identity injection, guiders and main denoising UNet for motion controlling. As for the four facial conditions, we randomly dropout either expression-related condition or shape-aware pose condition for their possibly repeated representation to make our network extract information from all conditions, instead of learning from easier one. The purpose of the first stage is to adapt the prior knowledge of Stable Diffusion into human face domain. 
During the second training stage, we only adjust the temporal layer to maintain the temporal consistency across frames. 
All the face images and frames are resized to $512\times 512$. We finetune the model for 260,000 steps using a batch size of 32. In the second training stage, we focus on training the temporal layer for 40,000 steps using 16-frame video sequences with a batch size of 16. We use same augmentation in both two stages for robust training. Apart from the reconstruction loss on the whole image from \Cref{eq:ldmloss}, we also add two losses on the whole facial area and several facial regions (e.g., eyes, ears, mouth and nose) to help network focus on the facial appearance and expression. The learning rate is $1\times10^{-5}$ in two stages. We fine-tuned the 3D face model~\cite{wang2024_3ddfav3} for higher accuracy. The temporal attention layers are initialized with pretrained model in AnimateDiff~\cite{guo2024animatediff}.
We optimize the overall framework using Adam on 4 NVIDIA H800 GPUs.

\subsection{Metrics}
Following common practice ~\cite{chen2020simswap, li2019faceshifter}, we randomly sample 10 images from each video to perform face swapping and compute the image-level metrics including ID retrieval, pose error, expression error, eye motion error and mouth position error. To compute ID Retrieval, we first extract the identity feature using a different face recognition model~\cite{wang2018cosface} from our face encoder. For each swapped face, we compute the nearest face from all the frames in FF++ using the cosine similarity and check whether it is from the source video. The pose error is computed by the mean L2 distance of pitch, yaw, and roll between the swapped face and the target face estimated by a different pose estimator~\cite{dad3dheads}. The expression error is the L2 distance between the expression vectors extracted by~\cite{dad3dheads} of the swapped face and the target face. Both eye motion error and mouth position error are evaluated by 2D keypoints to compare the position consistency between swapped faces and target faces. We also sample 30 frames to calculate video-level metrics containing CLIP frame consistency and warping error~\cite{lei2020blind}. For CLIP frame consistency, we use CLIP~\cite{radford2021clip} to obtain feature vectors of consecutive frames and then compute the cosine similarity between consecutive frames. To compute warping error, we calculate the warp differences on every two frames with an optical flow estimation network~\cite{ilg2017flownetv2} and calculate the final score using the average of all the pairs.
\begin{table}[t]

\centering

\setlength{\tabcolsep}{1mm}
\begin{tabular}{c|c|cc|cc}
\toprule
\multirow{1}{*}{Methods} & \multicolumn{1}{c|}{ID Retri. $\uparrow$} & \multicolumn{1}{l}{\multirow{1}{*}{Pose$\downarrow$}} & \multicolumn{1}{l|}{\multirow{1}{*}{Expr.$\downarrow$}} & \multirow{1}{*}{Mouth$\downarrow$} & \multirow{1}{*}{Eye$\downarrow$} \\ 
\midrule
Deepfakes               & 91.40 &3.32 &5.02 &7.11 &0.35                          \\
FaceShifter             & 96.00 &2.12 &3.49 &3.45 &0.29                      \\
MegaFS                  & 81.62 &5.33 &4.52 &9.34 &0.31 \\

SimSwap                  & 98.50 & \textbf{1.05} & \textbf{2.85} &2.39 &0.22 \\ 

DiffSwap                  & 17.21 &1.67 &3.05 &3.56 &0.24 \\

Face Adapter              & 98.69 & 1.90 &4.15 &4.08 &0.23 \\ 
\midrule
Ours & \textbf{99.20} & 1.73&3.08 & \textbf{1.69} & \textbf{0.16} \\

\bottomrule
\end{tabular}
\caption{Quantitative comparisons with state-of-the-art methods on FF++.
}
\label{tab:quantitative_comparison}

\end{table}
\subsection{Quantitative Comparisons}
Our method is compared with six methods including Deepfakes~\cite{deepfakes}, FaceShifter~\cite{li2019faceshifter}, 
MegaFS~\cite{zhu2021megafs},
SimSwap~\cite{chen2020simswap}, DiffSwap~\cite{zhao2023diffswap} and Face Adapter~\cite{han2024faceadapter}.
For Deepfakes and FaceShifter, we use their released face swapping results of the sampled 10,000 images. For MegaFS, SimSwap, DiffSwap, and Face-Adapter, the face swapping results are generated with their released codes. Table~\ref{tab:quantitative_comparison} shows that our method achieves the best scores under most evaluation metrics, including identity retrieval, mouth position error and eye motion error. These results validate the controllable superiority of our method.
Our pose error results are slightly poorer than others due to DynamicFace altering face shape, affecting the face shape-sensitive head pose estimator used.
Figure~\ref{fig:ffhq_comparison} illustrates that DynamicFace is competitive in preserving the background, motion of target face, and identity of source face well, especially for shape. Figure~\ref{fig:video_showcase} shows that our method could achieve promising consistency and generate controllable facial expression across frames. DynamicFace can generate faces with higher quality rather than GAN-based methods, and achieve better controllable ability than other Diffusion-based methods.

\subsection{Ablation study}
In order to further explore the effect of composable face conditions and other modules in DynamicFace, we conduct a ablation study on 1) four facial conditions 2) identity injection modules 3) temporal consistency module.

\noindent\textbf{Significance of composable face conditions.}
Four facial conditions aim to provide necessary guidance for controllable generation. Figure~\ref{fig:ablation_figure} shows that each condition plays its role in face swapping and each of them is necessary. Background could provide detailed environmental information, ensuring the generation outside facial area as an inpainting paradigm. Normal map is well-aligned with the shape of source face image and maintains the pose of target face frames, which guide precise shape and pose generation. UV texture map reflects the lighting of target face, making illumination consistent during face swapping. For more detailed ablation studies, please refer to Supplementary Materials.

\begin{table*}[t]
\centering

\setlength\tabcolsep{3pt}
\begin{tabular}{l|cc|cccc|ccc}

\toprule
& \multicolumn{2}{c|}{Identity Preservation} & \multicolumn{4}{c|}{Motion Fidelity} & \multicolumn{3}{c}{Video Quality} \\

\cline{2-3}
\cline{4-7}
\cline{8-10}
& ID Retri. $\uparrow$ & ID Simi. $\uparrow$ & Pose $\downarrow$ & Expr. $\downarrow$ & Mouth $\downarrow$ & Eye $\downarrow$ & Consistency $\uparrow$ & Warp. $\downarrow$ & Aesthetic $\uparrow$\\
\midrule

A) $w/o.$ motion module & \textbf{99.21} & \textbf{0.594} & $1.81$ & $3.12$ & \textbf{1.98} & $0.17$ & $95.78$ & $0.091$ & 3.26 \\
B) $w/.$ motion module & $98.90$ & $0.574$ &1.54 & \textbf{2.94} & $2.30$ & \textbf{0.16} & 99.02 & 0.046 & 3.48 \\
C) $w/.$ mm \& FusionTVO & $98.90$ & $0.574$ & \textbf{1.53} & \textbf{2.94} & $2.30$ & \textbf{0.16} & \textbf{99.04} & \textbf{0.045} & \textbf{3.51}\\
\bottomrule
\end{tabular}
\caption{Ablation study for the motion module and FusionTVO of our method during video face swapping.}
\label{tab:ablation_motion}

\end{table*}

\begin{table}[t]

\begin{center}
{
\setlength\tabcolsep{4pt}
\begin{tabular}{cc|ccc}
\toprule
Face Former & ReferenceNet & ID Simi. $\uparrow$  & Pose $\downarrow$ & Expr. $\downarrow$ \\ 
\midrule
 & \checkmark  &  $0.520$ & $1.35$& $3.23$ \\
\checkmark &  &  $0.515$ &$1.34$ &$3.18$  \\
\checkmark & \checkmark &\textbf{0.547} & \textbf{1.21} & \textbf{2.32} \\ 
\bottomrule
\end{tabular}
}
\end{center}

\caption{ Ablation study for the impact of Face Former and ReferenceNet on identity injection.
}
\label{tab:ablation-identity}

\end{table}
\noindent\textbf{Reliability of motion module.}
Despite our efforts to design precise conditions for generating swapped faces with identical motion to the target face, the intrinsic diversity of diffusion models leads to a noticeable variation.
It will generate faces with the same motion but slightly different details, making video face swapping unavailable. Here, we evaluate the importance of motion module from three aspects: identity preservation, motion fidelity, and video quality. Table~\ref{tab:ablation_motion} shows that DynamicFace could achieve great temporal consistency and aesthetic quality with FusionTVO.

\noindent\textbf{Effectiveness of identity injection modules.}
ReferenceNet and Face Former aim to enhance identity preservation through detailed-level and high-level feature injection, respectively. To ensure a fair comparison, we trained three models under identical settings to evaluate the impact of these identity modules. As demonstrated in Table~\ref{tab:ablation-identity}, both components significantly improve identity extraction. Specifically, ReferenceNet extracts fine-grained textures and selectively injects discriminative facial features into the main denoising UNet, while Face Former leverages high-level semantic embeddings derived from a pretrained face recognition model to reinforce identity coherence.

%% file: sec/5_conclusion.tex
\section{Conclusion}
We present a novel method named DynamicFace which leveraged the powerful pretrained diffusion model with delicately disentangled facial conditions, achieving promising face swapping results in both image and video domains. Experiments show that the designed facial conditions could give precise and unique evidence on required information (e.g., shape, expression, pose, lighting, and background). Several efforts have been taken to adapt the diffusion model to face swapping, including Face Former, ReferenceNet, Mixture-of-Guiders. Extensive experiments demonstrate that our framework can achieve superior results compared to previous methods, with better controllability and scalability. With above fine-grained facial conditions and FusionTVO, our method can achieve consistent video face swapping. We hope our attempt can inspire future work to further explore the formulation of face-swapping to achieve better results.

%% file: supp.tex


\section{Detailed Ablation Study on Facial Composable Conditions}

\begin{figure*}[t]
    \centering
    \includegraphics[width=\linewidth]{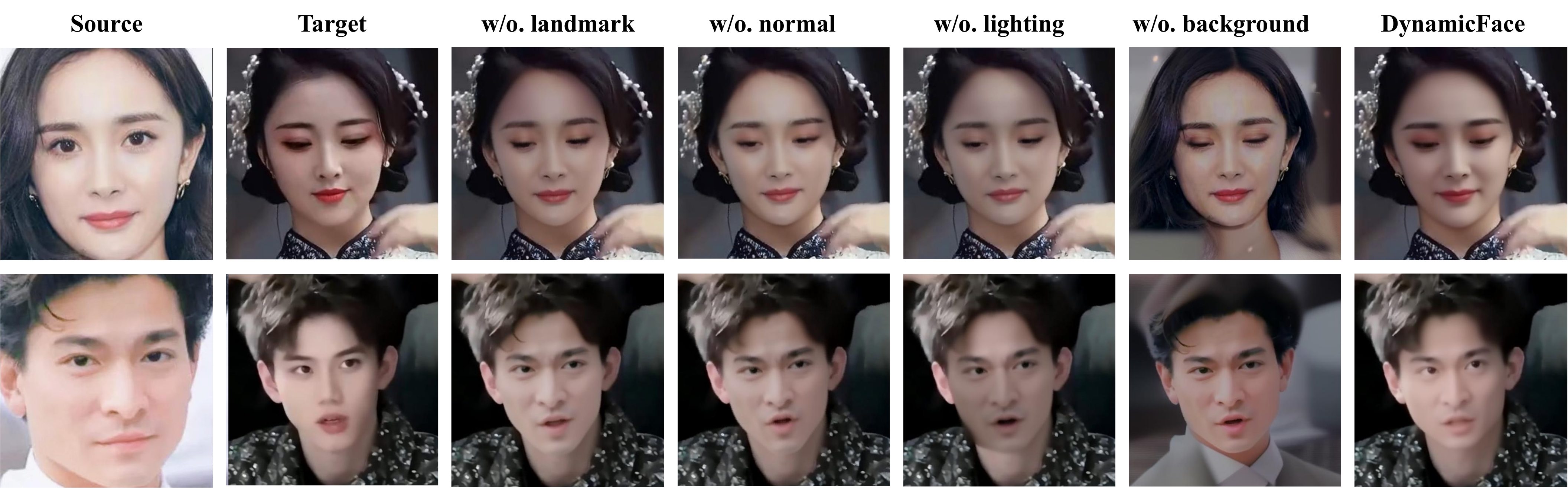}
    \caption{\textbf{Impact of disentangled facial conditions on face swapping.}}
    \label{fig:detail_ablation}
\end{figure*}

To rigorously validate the necessity of each facial condition, we conduct additional ablation experiments exclusively on the FaceForensics++ (FF++) dataset.  We systematically evaluated the impact of each composable facial condition by training the model while removing individual components on purely public dataset, ensuring reproducibility. Crucially, Table \ref{tab:detailed_ablation} validated that no single condition could be removed without significant performance degradation, confirming their complementary yet disentangled contributions. Figure \ref{fig:detail_ablation} shows the role of disentangled conditions and their collective effect.

\begin{table}[h]

\centering

\setlength{\tabcolsep}{1mm}
\begin{tabular}{l|c|cc|cc}
\toprule
\multirow{1}{*}{Methods} & \multicolumn{1}{c|}{ID Simi. $\uparrow$} & \multicolumn{1}{l}{\multirow{1}{*}{Pose$\downarrow$}} & \multicolumn{1}{l|}{\multirow{1}{*}{Expr.$\downarrow$}} & \multirow{1}{*}{Mouth$\downarrow$} & \multirow{1}{*}{Eye$\downarrow$} \\ 
\midrule
$w/o.$ landmark           & 55.45 &1.66 &3.46 &2.59 &0.47                        \\
$w/o.$ normal              & 56.32 &1.93 &3.13 &2.96 &0.43                      \\
$w/o.$ lighting              & 56.30 & \textbf{1.34} &2.89 &2.27 &0.46 \\

$w/o.$ bg                 & \textbf{73.05} & 2.43 & 4.01 &2.66 &0.48 \\ 

\midrule

Ours & 56.22 & 1.46 & \textbf{2.60} & \textbf{2.06} & \textbf{0.33} \\

\bottomrule
\end{tabular}
\caption{Ablation study on different facial conditions.
}
\label{tab:detailed_ablation}

\end{table}

\section{Why Disentanglement Matters}
\begin{figure}[t]
    \centering
    \includegraphics[width=\linewidth]{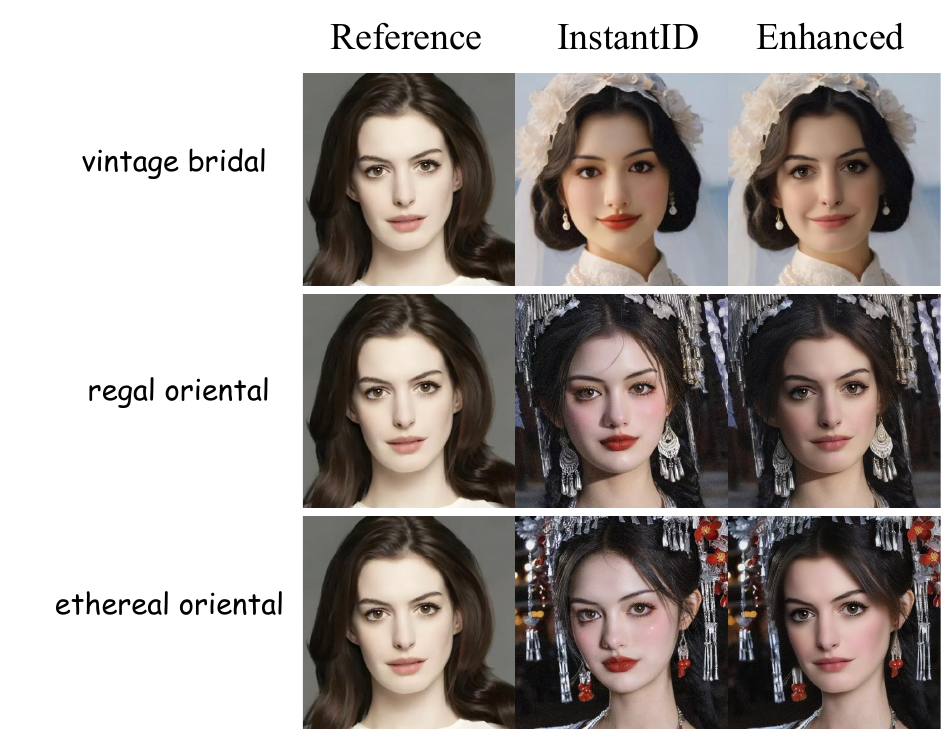}
    \caption{\textbf{Enhance results genearted by ID-Preserving works.}}
    \label{fig:instantid_enhance}
\end{figure}

The core strength of DynamicFace lies in its disentangled facial conditions, which enable independent control over identity, motion, and environmental attributes. We explore more possible applications to validate this: 

\noindent\textbf{Enhancing shape similarity of ID Preserving Text-to-Image Generation.}
Despite advancements in identity-preserving methods, existing approaches may still exhibit suboptimal ID similarity and inadequate face shape control. Shape-aware normal maps of DynamicFace anchor facial geometry to the source identity, enabling localized repairs (e.g. jaw realignment) without distorting the target’s expression or pose. We first generate portraits in different styles using ID Preserving methods and then enhance the generated results. Our method resolves identity leakage in the jaw and eye shape as shown in Figure \ref{fig:instantid_enhance}.

\begin{figure*}[h]
    \centering
    \includegraphics[width=0.95\linewidth]{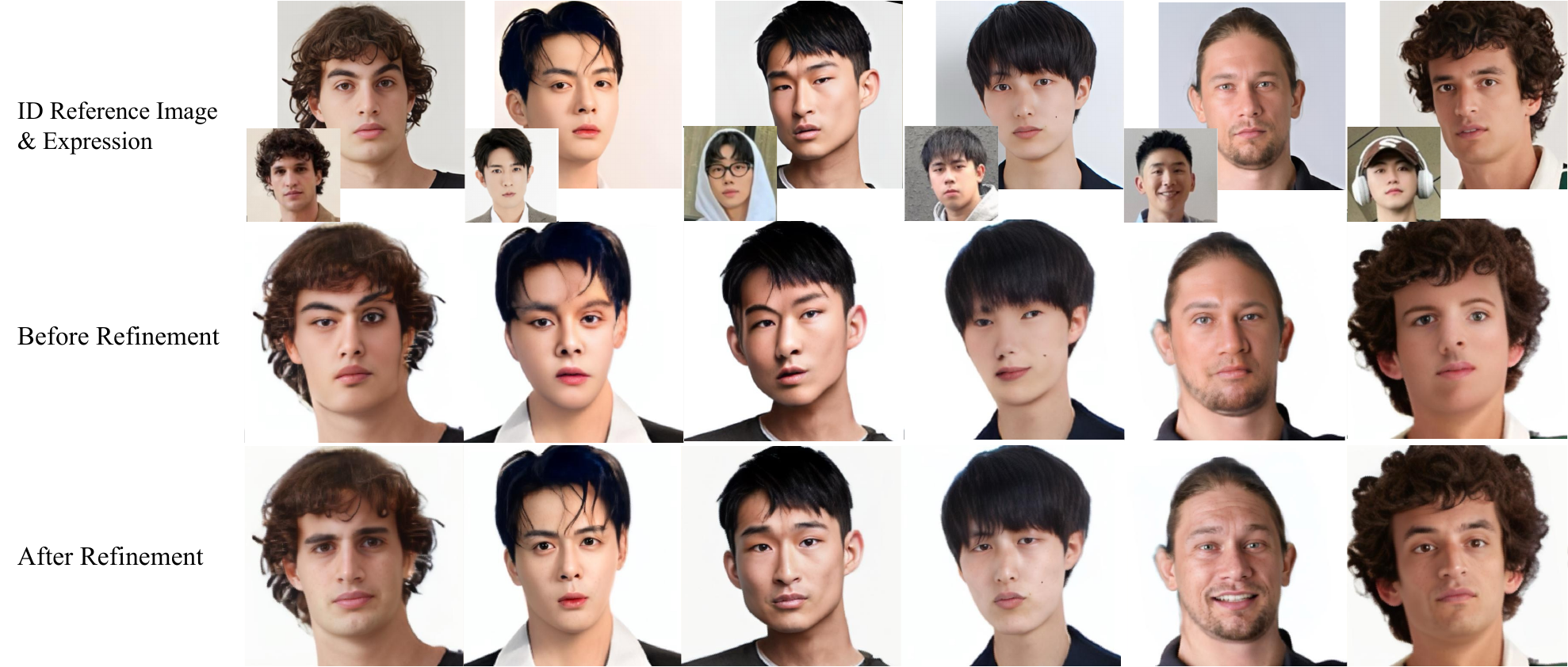}
    \caption{\textbf{Disentangled priors enable motion-consistent facial artifact correction in generated videos}}
    \label{fig:refine_man}
\end{figure*}

\begin{figure*}[h]
    \centering
    \includegraphics[width=0.95\linewidth]{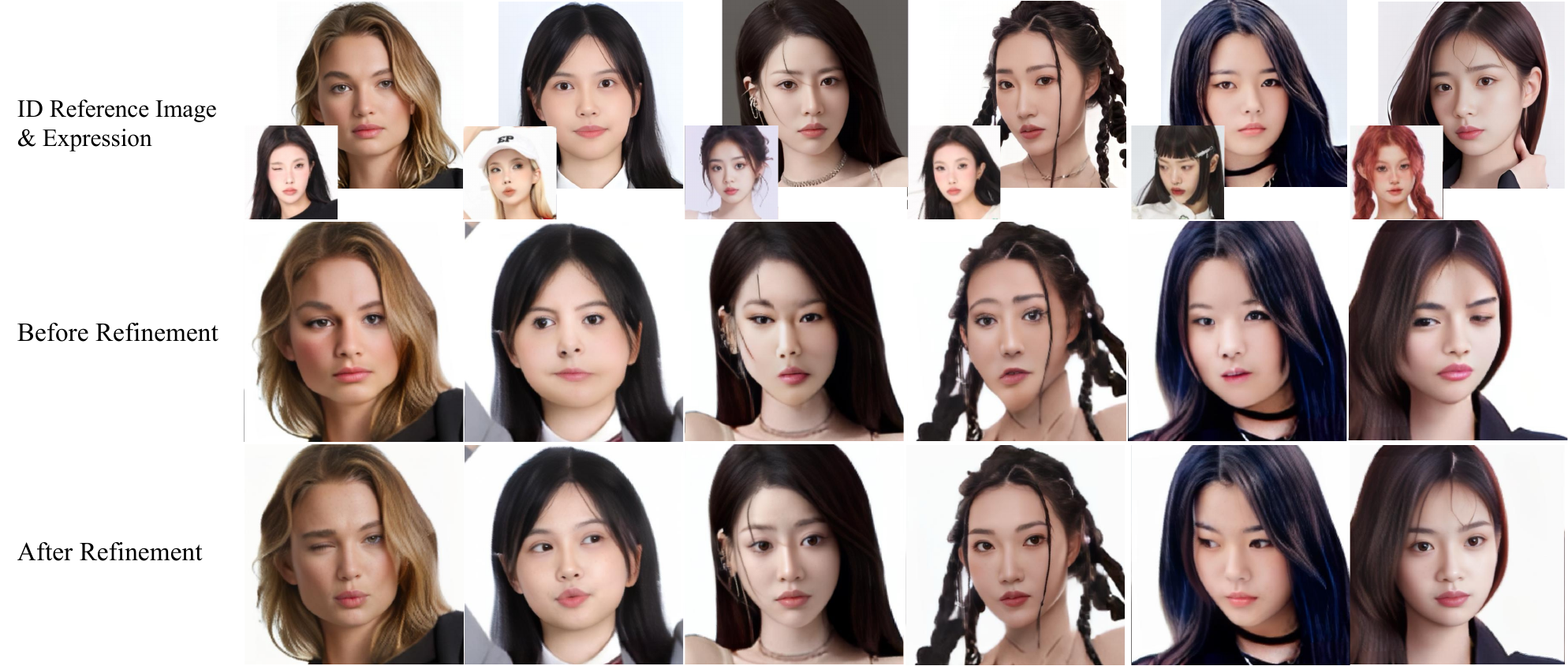}
    \caption{\textbf{Disentangled priors enable motion-consistent facial artifact correction in generated videos}}
    \label{fig:refine_woman}
\end{figure*}

\noindent\textbf{Motion-Consistent Artifact Restoring.}
By decoupling shape, expression, background, and illumination, DynamicFace could selectively restore corrupted regions without altering unrelated attributes (e.g., fixing a torn texture while preserving eye blinks). Our disentangled facial conditions enable precise control over motion attributes (e.g., expression, pose) while preserving the identity of the source face. This allows us to address facial artifacts in the results from other portrait generation works . Expression-related landmarks ensure repaired regions align with the original motion dynamics as shown in Figure \ref{fig:refine_man}, \ref{fig:refine_woman}.


\section{Extended Real World Visualizations}
We provide additional visual results on real-world videos to demonstrate robustness under challenging scenarios: extreme expressions and occlusions. The videos are sorted out together in the supplementary materials, including visual comparison with latest video face swapping methods.


\begin{figure*}[ht]
    \centering
    \includegraphics[width=\linewidth]{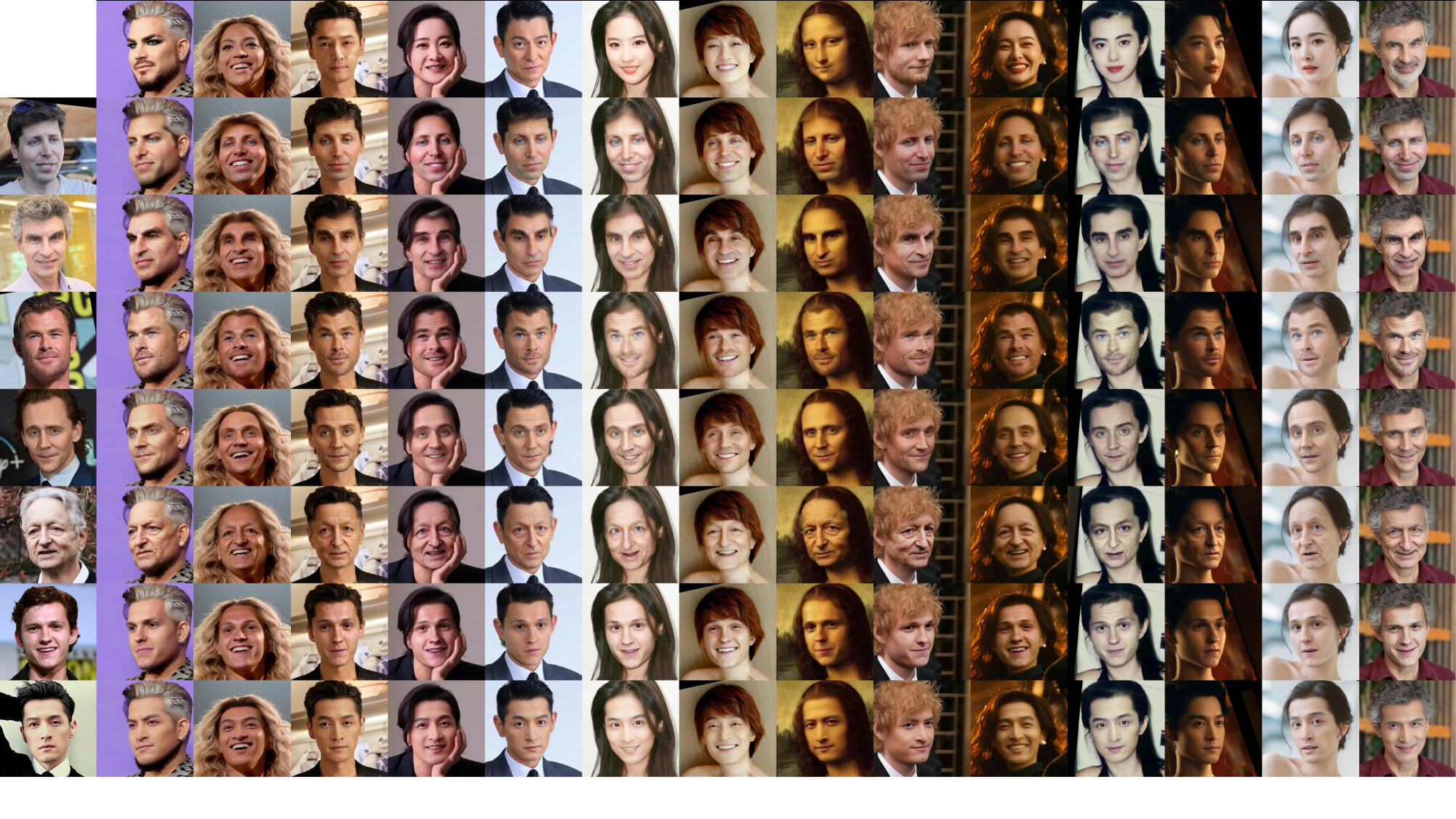}
    \caption{\textbf{Celebrity face swapping results under varied illumination and pose  conditions}}
    \label{fig:image_show2}
\end{figure*}

\begin{figure*}[ht]
    \centering
    \includegraphics[width=\linewidth]{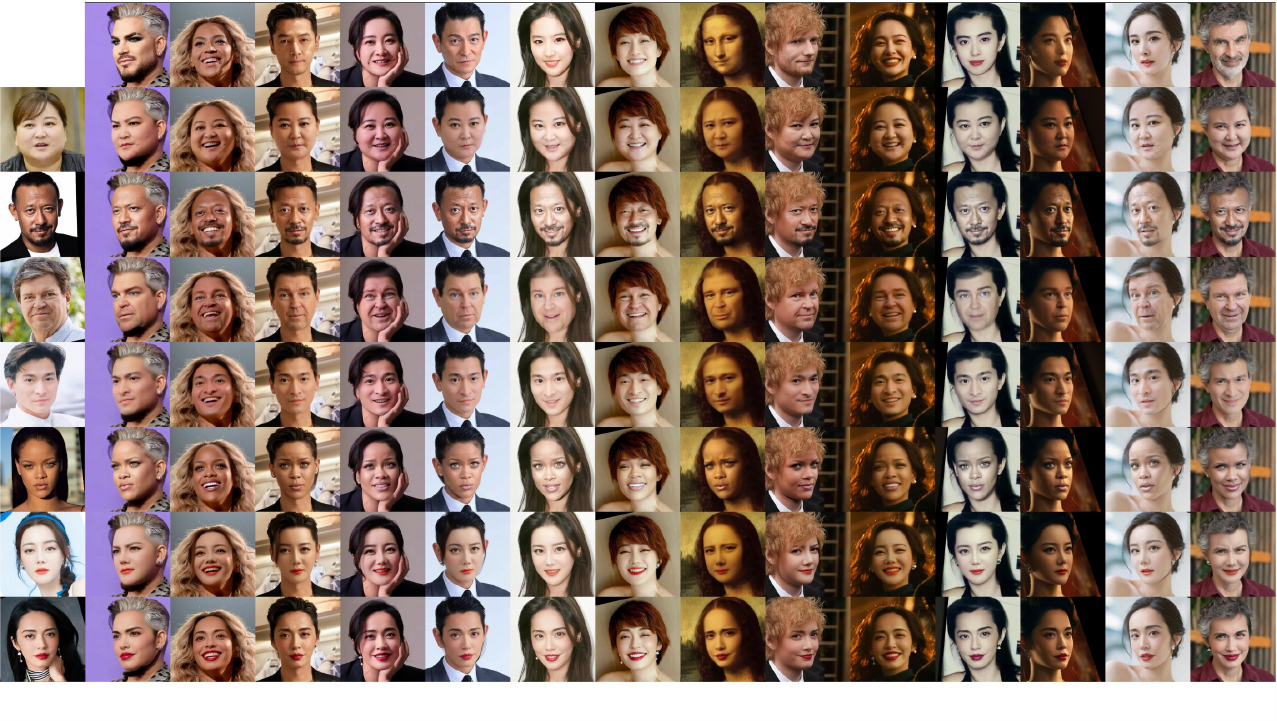}
    \caption{\textbf{Celebrity face swapping results under varied illumination and pose  conditions}}
    \label{fig:image_show3}
\end{figure*}